\documentclass[final,5p,times,twocolumn,authoryear]{elsarticle}

\usepackage{amssymb}
\usepackage{amsmath}
\usepackage{amsthm}
\usepackage{booktabs}
\usepackage{algorithm}
\usepackage{algpseudocode}
\usepackage{graphicx}
\usepackage{hyperref}
\usepackage{multirow}
\usepackage{makecell}
\usepackage{arydshln} 
\usepackage{float}
\usepackage{xcolor} 

\journal{Computer Networks and Communications}

\begin{document}

\begin{frontmatter}

\title{FederatedBytes: Tokenizer-Free, Byte-Level Small Language Models\\for Efficient Multimodal Federated Learning}

\author[uts]{Ali Sarabadani\corref{cor1}}
\ead{ali.sarabadani@student.uts.edu.au}

\author[uts]{Massimo Piccardi}
\author[uts]{Mehran Abolhasan}
\author[uts]{Farzad Tofigh}
\author[uts]{Ian Zhou}
\author[uts]{Justin Lipman}

\cortext[cor1]{Corresponding author}

\affiliation[uts]{
    organization={Cyber Digital Centre, University of Technology Sydney}
}

\begin{abstract}




Federated Learning (FL) is a powerful machine learning framework that allows building a global model from the contributions of multiple, distributed clients, while ensuring the privacy of their local data. However, its heavy computational and communication overheads challenge its use for contemporary AI workloads such as natural language processing (NLP) and multimodal analysis. For this reason, this paper presents a novel small language model that leverages byte-level text and multimodal data processing, enhanced by a linear attention mechanism, to enable manageable Federated Learning of NLP and multimodal analysis tasks. The proposed model, nicknamed  \text{FederatedBytes}, manages to process text and any other data types, including images, by treating its input as a mere sequence of bytes, while its lean linear attention mechanism allows abating the overall memory and computational footprint. To prove the effectiveness of the proposed model we have carried out an extensive performance evaluation on sentiment analysis tasks in multilingual and multimodal settings over the Multilingual Amazon Reviews and ScienceQA datasets. The experimental results show that FederatedBytes has been able to outperform comparable models such as ByT5 and LSTM, achieving higher accuracy while significantly reducing memory and communication overheads.
\end{abstract}

\begin{keyword}
Federated Learning \sep small language models \sep byte-level transformers \sep tokenizer-free \sep linear attention \sep multimodal classification.
\end{keyword}

\end{frontmatter}


\section{Introduction}

\noindent


Federated learning (FL) is a machine learning approach that allows training global machine learning models from multiple clients in a decentralized environment, while preserving the data privacy of the clients \cite{mcmahan2017communication}. This makes FL ideal for multi-client settings where both a global model and data privacy are required, such as in digital agriculture, healthcare, mobile communications, electric vehicles and many other use cases where data is sensitive, network bandwidth and computational resources are limited, and data sources are inherently distributed. For instance, a collaborative agricultural project might require training a global model for optimal harvesting by combining data from multiple farms, each with potentially different languages, data types (e.g., text reports, images and sensor readings), and low-resource edge devices.

Among contemporary AI workloads, natural language processing (NLP) plays a special role for its ability to automatically extract useful information from text written in a multitude of languages. The field of NLP has recently been revolutionized by the widespread adoption of language models such as ChatGPT, Gemini and Claude, that have been successfully applied to a range of tasks such as sentiment classification, machine translation, document summarization, question answering and reasoning. 



However, these models typically rely on a processing step known as \textit{tokenization}, which entails breaking the input text into predefined units roughly corresponding to individual words or part thereof. Tokenization is a substantial process requiring large, language-specific or multilingual vocabularies, struggling with out-of-vocabulary words, and hindering the use of non-textual  inputs such as images and videos. 
These challenges are particularly acute for edge devices where computational resources are intrinsically limited, and magnified by FL environments with their heavy communication requirements over networks with limited bandwidth.

To mitigate vocabulary issues, byte-level language models such as ByT5 \cite{xue2022byt5} and bGPT \cite{Shangda2024bgpt} have recently been proposed to operate directly on raw UTF-8 bytes, effectively removing the tokenizer and using a fixed vocabulary of only 256 entries. While this improves flexibility, the existing token-free models are not optimized for tight memory and computational constraints of edge devices in a decentralized, multimodal setting.
For this reason, in this paper we propose \textit{FederatedBytes} - a novel byte-level, small language model specifically designed for federated learning. By operating directly on raw UTF-8 bytes, FederatedBytes eliminates the need for tokenization,
naturally handling any language or data format that can be represented as a sequence of bytes. In addition, its architecture incorporates a linear attention mechanism, which abates the quadratic complexity of a standard Transformer (the core module of all language models), making it feasible to process long sequences on resource-constrained devices. Our work is motivated by the need for a versatile, efficient, and private model for multi-client industrial applications, where previous models such as ByT5 and bGPT prove inadequate due to their high resource consumption. To this aim, the novelty of FederatedBytes lies in its unique combination of tokenizer-free byte-level processing, a memory-efficient linear attention mechanism, small size, and seamless integration into a privacy-preserving federated learning framework. Overall, this paper makes a threefold contribution: 
\begin{itemize}
    \item We propose FederatedBytes, a byte-level, small language model tailored for federated NLP capable of handling multilingual and multimodal data on edge devices.
    \item We integrate into the model an efficient linear-attention mechanism that significantly lowers memory usage and extends the context window for processing long byte sequences - a critical requirement in many real-world applications.
    \item We perform ample experiments and comparisons on the Multilingual Amazon Reviews and ScienceQA multimodal datasets, showing that FederatedBytes has been able to achieve superior accuracy compared to an established byte-level model such as ByT5 and the LSTM, while substantially reducing training time, communication and computational overheads.

\end{itemize}


\section{Related Work}
Our work is positioned at the intersection of three active research areas: federated learning (FL) for natural language processing (NLP), byte-level text processing, and the development of efficient Transformer architectures. While significant progress has been made in each domain individually, FederatedBytes is the first model to merge these concepts into a unified framework designed for private, efficient, and multilingual NLP on decentralized data.

The main advantage of federated learning is that it aggregates locally-trained models without the need to centralize the raw data, and its baseline aggregation mechanism is Federated Averaging (FedAvg) \cite{mcmahan2017communication}. Early large-scale deployments such as Gboard's prediction of the next-word demonstrated the practicality of FL for NLP tasks on real devices \cite{hard2018federated}. More recently, frameworks such as FedNLP \cite{lin2021fednlp} 
have standardised comparisons of FL optimisers on Transformer-based tasks, highlighting the challenges that data heterogeneity poses for attention-based models such as Transformers. This has spurred research into personalised variants, such as FedTP \cite{zhang2022fedtp}, which adapts self-attention mechanisms in each client to counter the drift caused by non-IID (i.e., non independent and identically-distributed) data. However, these frameworks predominantly operate on tokenized text, inheriting the limitations of predefined vocabularies.

In parallel, byte-level models have emerged as a powerful alternative to token-based NLP approaches. By operating directly on raw byte sequences, models such as ByT5 \cite{xue2022byt5} and byte-level GPT (bGPT) \cite{Shangda2024bgpt} eliminate the need for tokenization, making them inherently 
multilingual, robust to out-of-vocabulary words and noise, and multimodal. Recent research by Meta AI into byte-level  Transformers \cite{Pagnoni2024blt} has further validated the scalability and potential of this approach. A critical gap, however, is that these powerful models have been designed and evaluated almost exclusively in centralized training settings, overlooking the challenges of deploying them in a privacy-preserving, federated context.

Addressing the computational demands of Transformers, especially on long byte sequences, has been another key research goal. The standard self-attention mechanism has a quadratic time and memory complexity with respect to the input length, which is a major bottleneck. This has led to the development of efficient, linearized attention mechanisms such as Linformer and Performer \cite{katharopoulos2020transformers}, and memory-efficient architectures such as the  Longformer \cite{beltagy2020longformer}. These innovations make it possible to efficiently process long inputs, a crucial capability for a byte-level model. 


For the reader's reference, Table~\ref{tab:related_work_paradigms} situates our work with respect to these key research threads, while Table~\ref{tab:related_work_models} provides a detailed comparison of specific token- and byte-level models.

\begin{table*}[htbp]
\centering
\caption{High-level comparison of key token- and byte-level models.}
\label{tab:related_work_paradigms}
\resizebox{\textwidth}{!}{%
\begin{tabular}{lcccc}
\toprule
\text{Model/Framework} & \text{Data Granularity} & \text{Learning Paradigm} & \text{Context Handling} & \text{Primary Contribution} \\
\midrule
FedNLP \cite{lin2021fednlp} & Token-Level & Federated & Standard ($O(L^2)$) & Benchmarking FL for token-based NLP. \\
ByT5 \citep{xue2022byt5} & Byte-Level & Centralized & Standard ($O(L^2)$) & State-of-the-art among early byte-level text processing. \\
Longformer \citep{beltagy2020longformer} & Token-Level & Centralized & Efficient (Linearized) & Efficiently processing long token sequences. \\
BLT \citep{Pagnoni2024blt} & Byte-Level & Centralized & Standard ($O(L^2)$) & Scaling byte-level models for large datasets. \\
\midrule
\text{FederatedBytes (Ours)} & \text{Byte-Level} & \text{Federated} & \text{Efficient (Linearized)} & \text{Unified framework for private, efficient, byte-level FL.} \\
\bottomrule
\end{tabular}%
}
\end{table*}

\begin{table*}[t!]
\centering
\caption{Detailed comparison of specific models and methods. This table analyzes the key innovations and suitability for federated learning of the models discussed. While powerful, centralized models such as ByT5 and BLT are not inherently designed for the communication and resource constraints of FL.}
\label{tab:related_work_models}
\small
\setlength{\tabcolsep}{4pt}
\renewcommand{\arraystretch}{1.05}
\resizebox{.9\textwidth}{!}{%
\begin{tabular}{l p{3.2cm} p{3.2cm} p{3.7cm}}
\toprule
\text{Model/Method} & \text{Key Innovation} & \text{Primary Use Case} & \text{Suitability for Federated Learning} \\
\midrule
FedAvg \cite{mcmahan2017communication} & Decentralized model aggregation & General-purpose FL & Foundational, but model-agnostic and can be inefficient for large models. \\
ByT5 \cite{xue2022byt5} & Encoder-decoder Transformer on byte streams & Multilingual NLP in centralized settings & Low. High computational and memory requirements; not optimized for communication efficiency. \\
bGPT \cite{Shangda2024bgpt} & Decoder-only Transformer on byte streams & Generative modeling on raw bytes & Low. Similar resource constraints to ByT5; large model size poses communication challenges. \\
BLT \citep{Pagnoni2024blt} & Scaling byte-level models to very large parameter counts & Foundational research in large byte-level models & Very Low. Designed for large-scale, centralized data centers, making it impractical for on-device FL. \\
\midrule
\text{FederatedBytes (Ours)} & Efficient attention on byte streams within a federated framework & Multilingual and multi-modal NLP on decentralized data & \text{High.} Explicitly designed to be resource-efficient, reduce communication, and operate without a central data store. \\
\bottomrule
\end{tabular}%
}
\end{table*}

\section{The FederatedBytes Architecture}
\label{sec:architecture}

Synthesizing the insights from the related work in Tables 1 and 2, we have identified a clear gap: existing frameworks effectively address federated learning, byte-level processing, or efficient attention in isolation, but no single architecture unifies all three. Recent byte-level models are too computationally heavy for federated edge devices, while existing efficient FL models rely on tokenization, sacrificing the robustness of raw byte processing. To bridge this gap, we propose \textbf{FederatedBytes}, a unified framework designed to bring the resilience of byte-level processing to the resource-constrained edge.

\begin{figure*}[t]
    \centering
    \includegraphics[width=\textwidth]{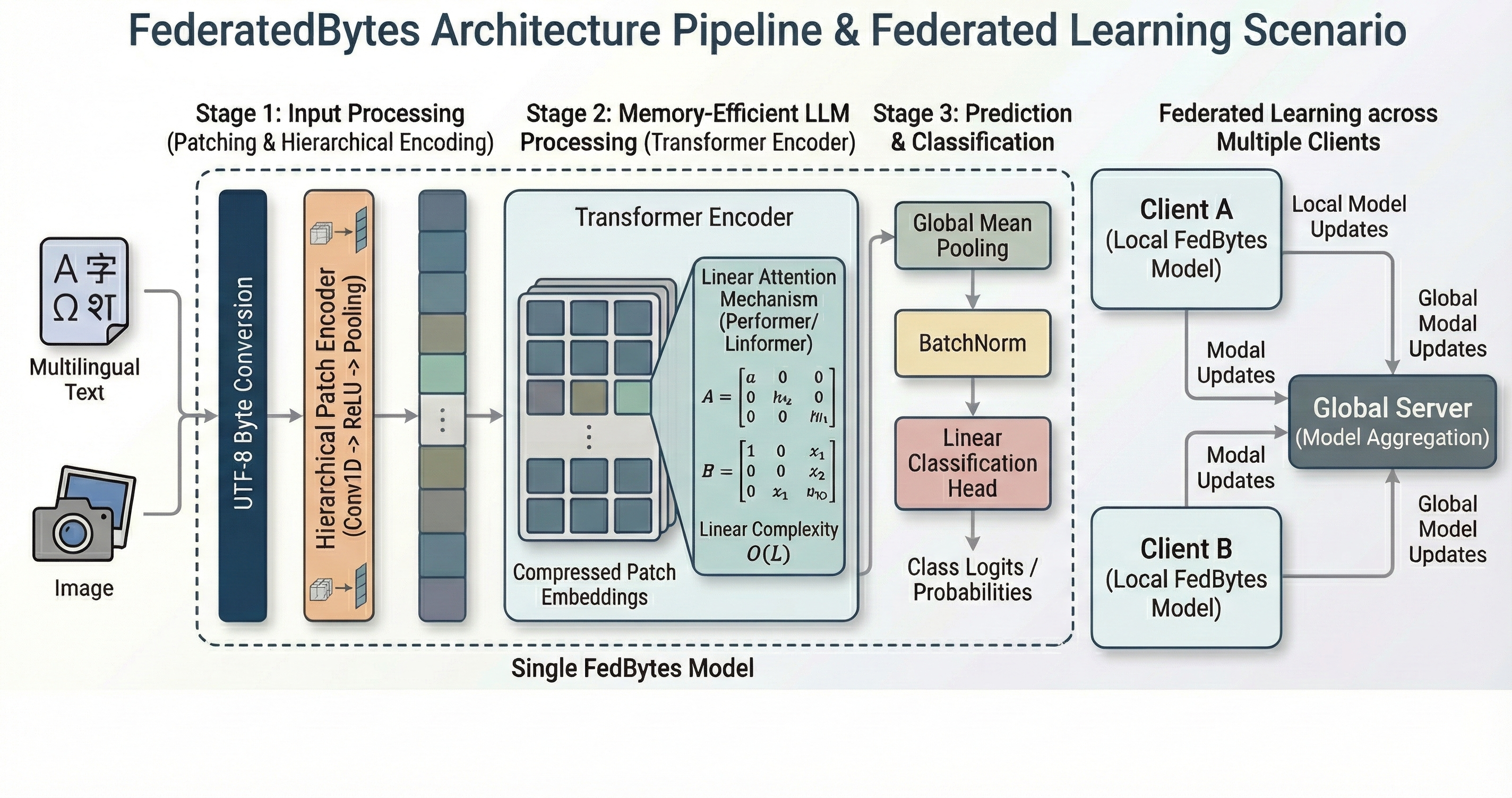}
    \caption{\textbf{End-to-end FederatedBytes Architecture.} The pipeline operates in three sequential stages tailored for efficient federated learning:
    \textbf{Stage 1 (Input Processing):} Raw multilingual text is encoded into UTF-8 bytes. These bytes are compressed into patch embeddings using a \textit{Hierarchical Patch Encoder} (1D Convolution $\rightarrow$ ReLU $\rightarrow$ Adaptive Pooling) to reduce sequence length.
    \textbf{Stage 2 (Memory-Efficient Processing):} The patch sequence is processed by a \textit{Transformer Encoder} utilizing the Performer mechanism. This approximates the full attention matrix using Orthogonal Random Features (ORFs) \cite{yu2016orthogonal}, achieving linear $O(L)$ time and space complexity.
    \textbf{Stage 3 (Prediction \& Classification):} The encoded representations are aggregated via global mean pooling, normalized via BatchNorm, and passed through a linear classification head to generate final class logits.}
    \label{fig:model_architecture}
\end{figure*}

In this section, we provide a high-level description of the architecture. As summarized in Figure \ref{fig:model_architecture}, the model operates directly on raw byte sequences and incorporates linear complexity attention mechanisms to handle large inputs efficiently. To align with the logical flow of the architecture, the pipeline is organized into three main sequential stages:

\begin{itemize}
    \item \textbf{Stage 1 (Input Processing: Patching and Hierarchical Encoding):} Conversion of multilingual text into UTF-8 byte streams, followed by segmentation and aggregation via a hierarchical convolutional encoder to reduce sequence length.
    \item \textbf{Stage 2 (Memory-Efficient LLM Processing):} Encoding of patch representations via a Transformer Encoder augmented with linear-map memory attention (Performer or Linformer) to optimize computational resources.
    \item \textbf{Stage 3 (Prediction \& Classification):} Aggregation of latent representations via global pooling followed by supervised classification.
\end{itemize}


\subsection{Input Processing: Patching and Hierarchical Encoding}
\label{sec:patching}

The first stage prepares raw text for the model by converting it into a compressed vector representation. Raw text inputs \(x\) are converted into byte sequences \(b = \{b_1, b_2, \dots, b_L\}\), where each \(b_i\) is a byte value corresponding to the UTF-8 encoding. Each byte \(b_i\) is then mapped to a continuous vector embedding \(e_i \in \mathbb{R}^d\), which will later be used for processing.

To process these sequences efficiently, we employ a \textbf{fixed patching} strategy augmented by a hierarchical encoder. The byte sequence is divided into patches of size $P_s$ (defaulting to 128). While simple averaging is supported, our primary configuration utilizes a \textbf{Hierarchical Patch Encoder}. This module applies a 1D Convolution over the patch dimension followed by a ReLU activation and an adaptive pooling layer to aggregate $P_s$ embeddings into one.

This step significantly optimizes computational resources. By reducing the effective sequence length from $L$ to $L/P_s$, the memory complexity is reduced from $O(L^2)$ to $O((L/P_s)^2)$ before the main Transformer layers process the data.

\subsection{Memory-Efficient LLM Processing}
\label{sec:model_arch}

The core of our framework is the \texttt{FederatedBytes}, a custom \textbf{Transformer Encoder} designed for memory efficiency. Unlike standard decoder-only architectures, our model utilizes bidirectional encoding to capture full sequence context (non-causal). The architecture is composed of modular components explicitly configured for the resource constraints of federated clients.

\subsubsection*{Core Components and Hyperparameters}
The model follows a standard Transformer topology. Based on our primary experimental configuration, the key hyperparameters are a hidden dimension $d = 768$, $12$ attention heads, a depth of $12$ layers, and a dropout rate of $p = 0.1$. The model’s structure, in order of execution, is:

\begin{itemize}
    \item \texttt{(embed)}: An Embedding layer mapping byte values to dense vectors ($d=768$).
    \item \texttt{(patch\_encoder)}: A hierarchical convolutional block that aggregates raw byte embeddings into patch vectors.
    \item \texttt{(pos\_emb)}: Learnable positional embeddings added to the patch vectors to capture sequence order.
    \item \texttt{(att)}: A stack of memory-efficient Transformer Encoder blocks. Each block utilizes linear attention mechanisms to perform contextual feature extraction without quadratic memory costs.
    \item \texttt{(pool)}: A global mean pooling operation that aggregates the sequence of hidden states into a single vector representation.
    \item \texttt{(bn)}: A BatchNorm1d layer applied to the pooled features for normalization.
    \item \texttt{(dp)}: A Dropout layer for regularization.
    \item \texttt{(clf)}: A Linear layer acting as the classification head.
\end{itemize}


\subsubsection*{Memory-Efficient Attention}
The computational core of the model is its attention mechanism. In a standard Transformer, the attention score is calculated as:

\begin{equation}
\text{Attention}(Q, K, V) = \text{softmax}\left(\frac{Q K^\top}{\sqrt{d_k}}\right) V
\end{equation}

where $Q, K \in \mathbb{R}^{L \times d}$ are the query and key matrices. The term $Q K^\top$ produces an $L \times L$ matrix. This standard self-attention presents a significant bottleneck for long sequences due to its quadratic space and time complexity, $O(L^2)$. For example, for our byte sequences where $L$ can reach 16,384, this matrix creates an imposing memory footprint.

To address this, we integrate the \textbf{Performer} mechanism \cite{choromanski2020rethinking}, which utilizes the \textit{kernel trick} to linearize attention. Instead of computing the massive $L \times L$ attention matrix, Performer approximates the softmax kernel using a random feature map $\phi(\cdot)$. This allows us to decompose the attention equation using the associative property of matrix multiplication:

\begin{equation}
\text{Attention}(Q, K, V) \approx \underbrace{\phi(Q)}_{L \times m} \left( \underbrace{\phi(K)^\top}_{m \times L} \underbrace{V}_{L \times d} \right)
\end{equation}

By changing the order of multiplication, we avoid constructing the $L \times L$ matrix entirely. Instead, we first compute $\phi(K)^\top V$, resulting in a small $m \times d$ summary matrix, and then multiply by $\phi(Q)$. The overall complexity drops from $O(L^2 d)$ to $O(Lmd)$, where $m$ is the number of random features ($m \ll L$).

\vspace{0.3cm}
\noindent\textbf{Matrix Decomposition Sample:}
To visualize this reduction, consider the dimensionality change during the forward pass:

\[
\text{Standard:} \quad 
\underbrace{\begin{bmatrix} 
    \dots & \dots \\ 
    \dots & \dots 
\end{bmatrix}}_{L \times L \text{ (Bottleneck)}} 
\times 
\begin{bmatrix} 
    \vdots \\ 
    V \\ 
    \vdots 
\end{bmatrix}
\]

\[
\text{Performer:} \quad 
\underbrace{\begin{bmatrix} 
    \vdots \\ 
    \phi(Q) \\ 
    \vdots 
\end{bmatrix}}_{L \times m} 
\times 
\underbrace{\left( 
    \begin{bmatrix} 
        \dots & \phi(K)^\top & \dots 
    \end{bmatrix} 
    \times 
    \begin{bmatrix} 
        \vdots \\ 
        V \\ 
        \vdots 
    \end{bmatrix} 
\right)}_{m \times d \text{ (Compact Context)}}
\]

Our implementation specifically employs Orthogonal Random Features (ORFs) \cite{yu2016orthogonal} to construct $\phi(\cdot)$. Unlike standard random Fourier features which are drawn independently, ORFs impose an orthogonality structure on the random projection weights. This theoretical guarantee reduces the variance of the approximation, providing a more stable estimate of the softmax kernel and encouraging consistent convergence during federated training rounds.



\subsection{Prediction and Classification}
\label{sec:prediction}

Once the attention mechanism has efficiently encoded the long byte sequence into a set of context-aware patch representations, the architecture must synthesize this information for the downstream task (i.e., classification in our case). In this final stage, the sequence of $N = L/P_s$ encoded features, $\mathbf{H} \in \mathbb{R}^{N \times d}$, is first compressed via global mean pooling to produce a single, fixed-size context vector $\mathbf{h}_{pooled}$.

To ensure training stability and prevent overfitting—crucial in federated settings where client data can be sparse—this pooled vector undergoes batch normalization and dropout before the final projection. The output probabilities $P \in \mathbb{R}^K$ for $K$ classes are then computed as:

\begin{equation}
P = \text{softmax}\left( \mathbf{W}_o \cdot \text{dropout}(\text{BN}(\mathbf{h}_{pooled})) + \mathbf{b}_o \right)
\end{equation}

where $\mathbf{W}_o$ and $\mathbf{b}_o$ denote the weights and biases of the final linear classification head. The entire network is trained end-to-end by minimizing the cross-entropy loss between $P$ and the ground-truth labels in the chosen training set.

\section{Methodology: Model, Data, and Training}
\label{sec:methodology}

Having defined the architectural components of FederatedBytes, we now detail the specific implementation and experimental framework used to evaluate its efficacy. Our approach realizes this architecture as a lightweight, byte-level Transformer Encoder optimized for efficient training in resource-constrained settings. The methodology encompasses three core components: (1) a hierarchical input representation that processes raw text as compressed byte patches; (2) a memory-efficient model implementation utilizing bidirectional attention with linear complexity; and (3) a rigorous federated training pipeline that transforms these patch representations into robust classification outputs.

\subsection{Data Pipeline with Worked Example}
\label{subsec:pipeline_example}

\begin{figure*}[t!]
    \centering
    \includegraphics[width=\textwidth]{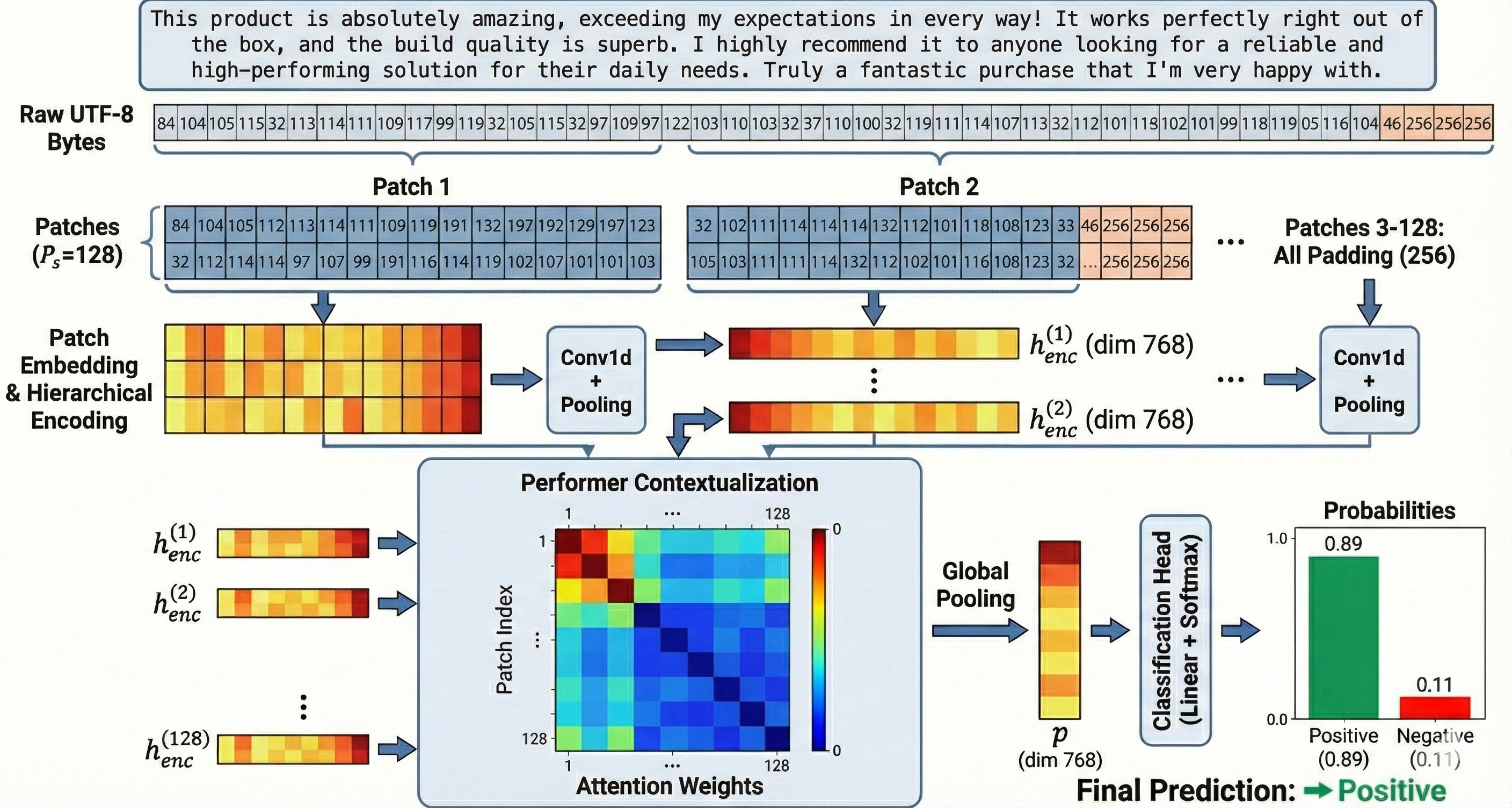}
    \caption{\textbf{The FederatedBytes Inference Pipeline.} An illustrative walkthrough of the running example (``This product is amazing...'') as it passes through the three stages of the model architecture. \textbf{Stage 1:} Input text is patched and processed by the Hierarchical Encoder (Conv1d) to reduce sequence length. \textbf{Stage 2:} Patches are contextualized via Performer linear attention. \textbf{Stage 3:} Features are pooled for final classification.}
    \label{fig:inference_flow}
\end{figure*}

To illustrate the pipeline, we use the running example of a short Amazon review:
\begin{quote}
    \textit{``This product is amazing and works perfectly!''}
\end{quote}
We trace the transformation of this input through the three architectural stages defined in Section 3.

\subsubsection{Stage 1: Input Processing (Patching and Averaging)}

\paragraph{Byte Encoding}
The raw text is first converted into a sequence of UTF-8 integers, one per character:
\[
\resizebox{0.95\columnwidth}{!}{%
$
\text{``This product...''} \rightarrow
\begin{bmatrix}
84 & 104 & 105 & 115 & 32 & 112 & 114 & 111 \\
100 & 117 & 99 & 116 & 32 & 105 & 115 & 32 \\
97 & 109 & 97 & 122 & 105 & 110 & 103 & 32 \\
97 & 110 & 100 & 32 & 119 & 111 & 114 & 107 \\
115 & 32 & 112 & 101 & 114 & 102 & 101 & 99 \\
116 & 108 & 121 & 33
\end{bmatrix}
$
}
\]
\textbf{Purpose:} To transform raw linguistic input into a numerical substrate suitable for arithmetic processing.





\paragraph{Padding and Patching}
The byte sequence of the review (44 bytes) is padded with the special token $\rho=256$ up to a fixed length ($L=16,384$ in our case). We then segment this sequence into 128 patches of size $P_s=128$ each. Unlike the previous small-patch example, the larger patch size in our architecture means that the entire 44-byte review fits within the first patch, $p_1$, filled to its full length by padding tokens:
\[ [\underbrace{84, 104, \dots, 119, 111, 114, 107}_{\text{44 text bytes}}, \underbrace{256, 256, \dots, 256}_{\text{84 padding tokens}}] \]

The second patch ($p_2$) and all subsequent patches comprise purely padding tokens.

\textbf{Purpose:} Standardize sequence length and create structured patch inputs for the Transformer. Padding ensures all inputs are of uniform length, while the use of token 256 prevents conflict with valid UTF-8 byte values (0–255).



\paragraph{Tabular View of Patch Content}
Since the model utilizes a patch size of $P_s=128$, the entire example text fits within the first patch. Table 5 visualizes the first 32 active bytes of \textbf{Patch 1}. The remaining 96 indices (32--127) of this patch are filled with the padding token $\rho=256$.

\begin{table}[h!]
\centering
\caption{Byte content of the first 32 positions in Patch 1.}
\resizebox{\columnwidth}{!}{
\begin{tabular}{|c|c|c|c|c|c|c|c|}
\hline
\textbf{T} (84) & \textbf{h} (104) & \textbf{i} (105) & \textbf{s} (115) & \textvisiblespace (32) & \textbf{p} (112) & \textbf{r} (114) & \textbf{o} (111) \\
\hline
\textbf{d} (100) & \textbf{u} (117) & \textbf{c} (99) & \textbf{t} (116) & \textvisiblespace (32) & \textbf{i} (105) & \textbf{s} (115) & \textvisiblespace (32) \\
\hline
\textbf{a} (97) & \textbf{m} (109) & \textbf{a} (97) & \textbf{z} (122) & \textbf{i} (105) & \textbf{n} (110) & \textbf{g} (103) & \textvisiblespace (32) \\
\hline
\textbf{a} (97) & \textbf{n} (110) & \textbf{d} (100) & \textvisiblespace (32) & \textbf{w} (119) & \textbf{o} (111) & \textbf{r} (114) & \textbf{k} (107) \\
\hline
\end{tabular}
}
\end{table}

\paragraph{Embedding and Hierarchical Encoding}
Once the input text is segmented into fixed-size byte patches, the next step is to project these raw integers into a continuous latent space. Each byte integer $b_i$ within a patch is first mapped to a dense embedding vector $e_i \in \mathbb{R}^{768}$.

However, processing every individual byte embedding would be computationally prohibitive. To reduce sequence length while retaining local context, we utilize a \textbf{Hierarchical Patch Encoder}. Unlike simple averaging, this module applies a learned transformation to the patch. Table \ref{tab:embeddings} illustrates the embeddings for the first few bytes of a patch before this aggregation takes place.

\begin{table}[h!]
\centering
\caption{Example embeddings (first 4 dimensions) for the first 8 bytes of Patch 1.}
\label{tab:embeddings}
\resizebox{0.8\columnwidth}{!}{%
    \begin{tabular}{ccccc}
        \toprule
        Byte & $e_{(1)}$ & $e_{(2)}$ & $e_{(3)}$ & $e_{(4)}$ \\
        \midrule
        84 (T)      & 0.12  & -0.08 & 0.33  & 0.44  \\
        104 (h)     & -0.21 & 0.15  & 0.05  & -0.36 \\
        105 (i)     & 0.19  & -0.11 & 0.29  & 0.07  \\
        115 (s)     & 0.03  & 0.48  & -0.22 & -0.15 \\
        32 (space)  & 0.00  & 0.00  & 0.00  & 0.00  \\
        112 (p)     & 0.44  & -0.05 & 0.19  & 0.09  \\
        114 (r)     & -0.09 & 0.25  & -0.16 & 0.38  \\
        111 (o)     & 0.28  & -0.14 & 0.07  & -0.21 \\
        \bottomrule
    \end{tabular}
}
\end{table}

We aggregate the sequence of embeddings within the patch ($128 \times 768 \rightarrow 1 \times 768$) using a 1D Convolution followed by adaptive pooling. Formally, for the $k$-th patch matrix $E^{(k)}$, the encoded representation $h^{(k)}$ is computed as:

\begin{equation}
    h^{(k)} = \text{Pool}\left(\sigma\left(\text{Conv1d}(E^{(k)})\right)\right)
\end{equation}

where $\sigma$ is the ReLU activation function. For this example, the resulting encoded vector begins:
\[
h^{(k)} = [0.07, 0.05, 0.08, 0.03, \dots]
\]

\textbf{Purpose:} To reducing sequence dimensionality and summarize local information within each patch via learned feature extraction, enabling efficient processing of contexts up to 16,384 bytes in size.

\subsubsection{Stage 2: Memory-Efficient Processing}

\paragraph{Contextualization}
Having transformed the raw byte stream into a sequence of dense patch representations, the architecture now shifts to modeling the relationships between them. The sequence of patch vectors is processed by the \textbf{Performer} attention mechanism. This efficient linear attention allows the model to learn dependencies between distant parts of the byte stream—for instance, relating the patch containing the sentiment-bearing word ``amazing'' to the entity ``product'' appearing much earlier in the sequence.

Table \ref{tab:attention} illustrates hypothetical attention weights between the first three patches, demonstrating how the model attends to relevant context regardless of distance:

\begin{table}[h!]
\centering
\caption{Hypothetical attention distribution across the first 3 patches.}
\label{tab:attention}
\resizebox{0.65\columnwidth}{!}{%
    \begin{tabular}{cccc}
        \toprule
         & Patch 1 & Patch 2 & Patch 3 \\
        \midrule
        Patch 1 & 0.60 & 0.25 & 0.15 \\
        Patch 2 & 0.20 & 0.65 & 0.15 \\
        Patch 3 & 0.10 & 0.30 & 0.60 \\
        \bottomrule
    \end{tabular}
}
\end{table}

\textbf{Purpose:} To capture long-range semantic relationships and global context across the various byte patches without incurring the quadratic cost of standard attention.

\subsubsection{Stage 3: Prediction \& Classification}

\paragraph{Pooling and Classification}
Once the sequence has been contextualized, the final stage is to synthesize this information into a decision. The encoded features $\mathbf{H} \in \mathbb{R}^{128 \times 768}$ are aggregated via \textbf{global average pooling} across the sequence dimension, reducing the input to a single global context vector of size 768.

\[
p = \text{mean}(\mathbf{H}) = [0.12, -0.05, 0.07, 0.19, \dots] \in \mathbb{R}^{768}
\]

This vector is then regularized and projected to the output space. It passes through a Batch Normalization layer, a Dropout layer, and finally the linear classifier to produce raw logits for the target classes (two in this case, as an example of binary classification):

\[
\hat{y} = \texttt{model.clf}(p) \quad \rightarrow \quad [1.45, -0.62]
\]

Applying the softmax function yields the final class probabilities:
\[
P(\text{positive}) \approx 0.888, \quad P(\text{negative}) \approx 0.112
\]
Consequently, the model correctly predicts the review sentiment as \textbf{positive}.

\textbf{Purpose:} To aggregate distributed information across all patches into a fixed-size representation and generate the final predictive probability scores.
\subsection{Training Process}

With the complete forward propagation pipeline—from raw byte patching to final classification—fully established, we now turn to the optimization procedure used to learn the model weights.

The model is trained using a supervised learning regimen designed to minimize the error between predicted logits and ground-truth labels. This process involves iterating over the training dataset in batches, computing the cross-entropy loss, and updating the model's approximately 30 million and 100 million parameters in FederatedBytes-30m and FederatedBytes-100m variants, respectively, via backpropagation. To ensure training stability and convergence, we utilize the proven AdamW~\cite{loshchilov2019} optimizer with weight decay and mixed-precision gradients.

In order to minimize the loss, the training process aims to maximize the probability assigned to the ground-truth class label for all samples in the training set (ideally approaching values of one). The following steps detail the flow of a single training step, using our running example of the positive Amazon review to illustrate the transformation of data.

\paragraph{Step 1: Batch Preparation (Simulation Scaling)}
To facilitate the large-scale simulation of 100 federated clients on a single NVIDIA A100 GPU, we adopt a strict memory-constrained training regimen. The training data is loaded in micro-batches of size $B=2$. Due to the memory footprint of the Performer attention map on large inputs ($L=16,384$), this micro-batching is essential to prevent Out-Of-Memory (OOM) errors while maintaining high client concurrency. We utilize gradient accumulation (128 steps) to achieve an effective batch size of $2 \times 128 = 256$. We assume our example is the first sample in the batch (index = 0) and has ground-truth label $y_0 = 1$ (positive class in a binary set). Table \ref{tab:train-input} summarizes this sample.

\begin{table}[h!]
\centering
\caption{Sample input data for our example review within a training micro-batch.}
\label{tab:train-input}
\resizebox{\columnwidth}{!}{
\begin{tabular}{|l|l|}
\hline
\text{Data Element} & \text{Sample Values for Review ($i=0$)} \\
\hline
Input Patches ($X_0$) & First patch: $[84, 104, 105, 115, 32, 112, 114, 111, \dots, 256]$ (128 integers) \\
Label ($y_0$) & $1$ \\
\hline
\end{tabular}
}
\end{table}

\textbf{Purpose:} To break the data into manageable pieces, allowing the simulation to fit 100 distinct client models within the VRAM constraints of a single A100 GPU.

\paragraph{Step 2: Forward Pass and Logit Generation}
The batch is then fed into the model. Table \ref{tab:train-forward-fixed} shows the tensor for our specific review sample as it is transformed through each layer of the \texttt{FederatedBytes}. Note the dimension's increase to 768 (hidden size) and the sequence length of 128 patches.

\begin{table*}[h!]
\centering
\caption{Tensor transformation for one sample during the forward pass.}
\label{tab:train-forward-fixed}
\resizebox{0.85\linewidth}{!}{
\begin{tabular}{l|l|l}
\hline
\text{Stage of Processing} & \text{Data Shape} & \text{Sample Value / Representation (first 4 dims shown)} \\
\hline
\text{1. Input Patch} & `(128,)` & `[84, 104, 105, 115, ...]` \quad (A single patch of 128 byte integers) \\
\hline
\text{2. Embedding} & `(128, 768)` & $
\begin{pmatrix}
0.12 & -0.08 & 0.33 & \dots \\
-0.21 & 0.15 & 0.05 & \dots \\
\vdots & \vdots & \vdots & \ddots
\end{pmatrix}
$ \quad (Matrix of 128 embedding vectors) \\
\hline
\text{3. Hierarchical Encoding} & `(768,)` & `[0.07, 0.05, 0.08, 0.03, \dots]` \quad (Encoded vector for the patch via Conv1d) \\
\hline
\text{4. Attention} & `(128, 768)` & `[0.09, -0.01, 0.15, 0.02, \dots]` \quad (Sequence of 128 contextualized vectors) \\
\hline
\text{5. Global Pooling} & `(768,)` & `[0.12, -0.05, 0.07, 0.19, \dots]` \quad (Mean of all 128 patch vectors) \\
\hline
\text{6. Classification} & `(2,)` & `[-0.62, 1.45]` \quad (Final 2-dim output logits for the sample) \\
\hline
\end{tabular}
}
\end{table*}

\textbf{Purpose:} To compute the probability of all classes in our class set by passing the input through the model's layers, transforming raw bytes into a final logit vector that represents the classification score for each class.

\paragraph{Step 3: Loss Computation}
The discrepancy between the model's logits and the ground-truth label is quantified using the standard Cross-Entropy Loss. For our example, the model produced logits $[-0.62, 1.45]$, and the true label is $1$. Therefore:
\[
\text{Probabilities} = \text{softmax}([-0.62, 1.45]) \approx [0.112, 0.888]
\]
\[
\mathcal{L}_0 = -\log(\text{Probability of true class}) = -\log(0.888) \approx 0.119
\]
The final loss is averaged across the batch.

\textbf{Purpose:} To calculate a single scalar value that measures the model's error on the sample, providing a differentiable objective signal for optimization.

\paragraph{Step 4: Backpropagation and Weight Update}
The batch loss is used to compute gradients for all trainable parameters via \texttt{loss.backward()}. Because the physical batch size ($B=2$) is restricted by the A100 simulation constraints, we perform gradient accumulation for 128 steps before the \texttt{AdamW} optimizer updates the model's weights.

\textbf{Purpose:} To decouple the logical batch size from physical memory limitations, ensuring stable convergence for the 100-client simulation despite the small micro-batch size.

\subsection{Evaluation Protocol}
To assess the model's generalization capabilities on unseen data, a comprehensive evaluation is performed on the test set of each dataset. The process mirrors the training forward pass but does not perform any weight updates.

\paragraph{Step 1: Inference Mode and Batch Processing}
The model is set to inference mode (\texttt{model.eval()}) and gradient calculations are disabled. The evaluation data, including our example review, is processed in batches. (For simplicity, in Table \ref{tab:eval-input} we show the same example used to illustrate the training, but in reality the training and test sets are always disjointed.)

\begin{table}[h!]
\centering
\caption{Sample input data for our example review within an evaluation batch.}
\label{tab:eval-input}
\resizebox{\columnwidth}{!}{
\begin{tabular}{|l|l|}
\hline
\text{Data Element} & \text{Sample Values for Review ($i=0$)} \\
\hline
Input Patches ($X_0$) & First patch: $[84, 104, 105, 115, 32, \dots]$ \\
Label ($y_0$) & $1$ \\
\hline
\end{tabular}
}
\end{table}

\textbf{Purpose:} To ensure deterministic model outputs by disabling the Dropout stochastic layer, and skip the unnecessary gradient computation.

\paragraph{Step 2: Prediction Generation}
A forward pass, identical to the one used in training, generates the logits. The final prediction is determined by applying an argmax operator. For reference, let's assume that the model, after training, produces slightly different logits for our sample:

\begin{table}[h!]
\centering
\caption{Prediction generation for our review sample during evaluation.}
\label{tab:eval-predict}
\resizebox{\columnwidth}{!}{
\begin{tabular}{|l|l|}
\hline
\text{Stage} & \text{Sample Value} \\
\hline
Final Pooled Vector & $[0.15, -0.09, 0.11, 0.23, \dots]$ \\
Generated Logits ($\hat{y}_{\text{logits}, 0}$) & $[-1.15, 2.50]$ \\
Probabilities & $\text{softmax}([-1.15, 2.50]) = [0.025, 0.975]$ \\
Prediction ($\hat{y}_{\text{pred}, 0}$) & $\text{argmax}([-1.15, 2.50]) = 1$ \quad (Positive) \\
\hline
\end{tabular}
}
\end{table}

\textbf{Purpose:} To obtain the model's final, discrete predictions and class probabilities for each sample in the unseen test set.

\paragraph{Step 3: Dataset-Wide Result Aggregation}
The prediction, true label, and logits for our sample are appended to lists that collect the results for the entire evaluation dataset (i.e., the test set) of size $N_{\text{eval}}$. Table \ref{tab:eval-aggregate} illustrates the lists.

\begin{table}[h!]
\centering
\caption{Appending our example's results to the aggregated evaluation lists.}
\label{tab:eval-aggregate}
\resizebox{\columnwidth}{!}{
\begin{tabular}{|l|l|}
\hline
\text{Aggregated List} & \text{State After Processing Our Sample} \\
\hline
All Labels ($y_{\text{all}}$) & $[1, \dots]$ \\
All Predictions ($\hat{y}_{\text{all}}$) & $[1, \dots]$ \\
All Logits ($\hat{y}_{\text{logits, all}}$) & $[ [-1.15, 2.50], \dots]$ \\
\hline
\end{tabular}
}
\end{table}

\textbf{Purpose:} To gather all necessary outputs to perform a comprehensive, dataset-wide performance analysis, which is far more reliable than single-batch metrics.

\paragraph{Step 4: Comprehensive Metric Calculation}
Using the complete, aggregated lists, a suite of metrics is calculated. Our example, having been correctly classified ($y_0=1, \hat{y}_{\text{pred}, 0}=1$), contributes positively to the overall accuracy.
\begin{itemize}
    \item \text{Accuracy:} The total number of matches between $y_{\text{all}}$ and $\hat{y}_{\text{all}}$, divided by $N_{\text{eval}}$.
    \item \text{Weighted F1-Score:} Calculated from $y_{\text{all}}$ and $\hat{y}_{\text{all}}$ to account for class imbalance.
    \item \text{AUC:} Calculated by comparing the true labels $y_{\text{all}}$ against the probabilities derived from $\hat{y}_{\text{logits, all}}$.
\end{itemize}

\textbf{Purpose:} To provide a robust quantitative assessment of the model's performance, covering its correctness (Accuracy), balanced performance (F1-Score), and discriminative power (AUC) on unseen data.







\subsection{Federated Learning Framework}
\label{sec:fl_framework}

Having established the training routine and evaluation metrics for a single FederatedBytes model, in this section we describe how multiple such models are orchestrated across a decentralized network to form the complete training pipeline.

To implement and simulate the federated learning process, we utilize the popular \textbf{Flower} framework \cite{flower}. Unlike centralized training, where data is aggregated, our framework distributes the models to the data. The workflow consists of three iterative steps:

\paragraph{1. Global Initialization}
A global model instance with parameters $\theta^0$ is initialized on the central server. In our simulation, the server distributes this initial state to a subset of available clients $C_r$ selected for the current round $r$.

\paragraph{2. Local Training}
Each selected client $k$ retrieves the global parameters $\theta^t$ and trains a local instance of the \texttt{FederatedBytes} on its private partition of the dataset $\mathcal{D}_k$. The client performs $E$ local epochs (set to 5 in our experiments) to minimize the cross-entropy loss:
\begin{equation}
\theta_k^{t+1} \leftarrow \text{LocalUpdate}(\theta^t, \mathcal{D}_k)
\end{equation}
This step utilizes the same byte-level patching and Performer attention mechanisms described in Section \ref{sec:architecture}.

\paragraph{3. Federated Averaging (Model Aggregation)}
Once local training is complete, clients upload their updated model parameters $\theta_k^{t+1}$ to the server. The server aggregates these updates using the Federated Averaging (FedAvg) strategy \cite{mcmahan2017communication}. The new global model $\theta^{t+1}$ is computed as the weighted average of the client models:
\begin{equation}
\theta^{t+1} = \sum_{k=1}^{K} \frac{N_k}{N} \theta_k^{t+1}
\end{equation}
where $N_k$ is the number of samples on client $k$, and $N = \sum N_k$ is the total number of samples across participating clients.
\paragraph{4. Simulation Environment and Scalability Analysis}
To validate this architecture at scale, we employ the \texttt{flwr[simulation]} engine backed by Ray\footnote{https://github.com/ray-project/ray}. We adopt a progressive scaling strategy, systematically increasing the client cohort size $K \in \{5, 20, 50, 100\}$ to record convergence behavior and system stability under varying levels of concurrency. This setup enables the simulation of up to 100 distinct clients with heterogeneous data partitions on a single NVIDIA A100 GPU, utilizing dynamic resource management to efficiently schedule and process concurrent client workloads.

\subsection{Security and Privacy Considerations}

Security and privacy are foundational to FederatedBytes' architecture. The framework implements a multi-layered, defense-in-depth strategy to protect data and model integrity, adhering to the core principles of privacy-preserving machine learning:

\begin{itemize}
    \item Data Privacy by Design: The cornerstone of the framework's privacy model is data localization. Raw, sensitive data never leaves the user's device. Only abstract model parameters (e.g., weights or gradients), which are learned from the local data, are shared for aggregation. This approach inherently prevents data leakage from a central server and aligns with modern data protection regulations such as the General Data Protection Regulation (GDPR).

    \item Secure Aggregation: While individual model updates do not contain raw data, they can still inadvertently leak information about the training data of a specific device. To mitigate this risk, FederatedBytes is designed to use secure aggregation protocols \cite{bonawitz2017practical}. These cryptographic methods allow the central server to compute the sum of all model updates from participating devices without being able to inspect any single, unencrypted update. This ensures that an honest-but-curious server cannot reverse-engineer a specific user's contribution.

    \item Differential Privacy: To provide formal, mathematical privacy guarantees against powerful inference attacks, client-side differential privacy can be applied \cite{dwork2006calibrating}. Before an update is sent, a carefully calibrated amount of statistical noise is added to the model parameters. This makes it computationally infeasible for an adversary to determine whether any particular individual's data was part of the training process, thus protecting against threats like membership inference attacks.

    \item Secure Communication Channels: All communication between client devices and the aggregation server is encrypted using Transport Layer Security (TLS). This secures the data in transit, ensuring both the confidentiality and integrity of the model updates. It provides a crucial line of defense against man-in-the-middle (MITM) attacks, eavesdropping, and data tampering during transmission.

    \item Robustness to Adversarial Attacks: Beyond privacy, security also involves ensuring the integrity of the final model. FederatedBytes can incorporate robust aggregation algorithms (e.g., using medians or trimmed means instead of a simple average) to defend against model poisoning attacks, where malicious clients attempt to corrupt the global model by sending faulty updates.
\end{itemize}

\section{Experimental Setup}
\label{sec:setup}

We have conducted a series of experiments to evaluate the performance of our proposed byte-level model. We have experimented with two variants, FederatedBytes-Base, which adopts a standard attention mechanism, and FederatedBytes-Performer, which adopts a linear attention, and two different sizes, nicknamed FederatedBytes-30M and FederatedBytes-100M with reference to their approximate parameter size. We have compared our models against established token-level and byte-level baselines of relatively small size (< 500M parameters): 1) GPT-2 Medium \cite{radford2019language}; 2) ByT5 \cite{xue2022byt5}; and 3) a 2-layer LSTM \cite{hochreiter1997lstm}. We have first evaluated all the models in a centralized training setting (a single model trained on the entire training set), and then benchmarked the top-performing models in a simulated federated learning (FL) environment.

\subsection{Datasets}
\label{sec:data}
We have used two diverse datasets to evaluate model performance in different languages and modalities:

\begin{table*}[t] 
\centering
\renewcommand{\arraystretch}{1.2}
\caption{Overview of the Datasets Used in the Experiments.}
\label{tab:dataset_cards}

\small 
\begin{tabular}{p{0.15\textwidth} p{0.32\textwidth} p{0.26\textwidth} p{0.20\textwidth}}
\toprule
\textbf{Property} & \textbf{Multilingual Amazon Reviews} & \textbf{ScienceQA} \\
\midrule
Task & Binarized 5-Class Sentiment Analysis & Multimodal Question Answering \\
Domain & Product Reviews & Scientific Questions \\
Classes & 5 (1-5 Stars) & Multiple Choice (varies) \\
Size & 200k samples per language & 21k samples \\
Languages & English, German, Chinese & English \\
Modality & Text & Text + Image \\
Source & \cite{prettenhofer2010cross} & \cite{lu2022learn} \\
\midrule
Truncation & \multicolumn{3}{p{0.78\textwidth}}{Note that, for Amazon Reviews, capping the input sequence length to 1024 bytes would truncate approx. 4\% of the samples, while a 2048-byte cap would limit truncation to $<1\%$ of the samples. Therefore, such cappings would have a limited impact on the measured performance.} \\
\bottomrule
\end{tabular}
\end{table*}

\begin{itemize}
    \item \text{Amazon Multilingual Reviews:} A sentiment classification dataset. We have used the subsets for Amazon (en), Amazon (de), and Chinese (zh). 
    
    
    The task is binary classification, where reviews with 4 or 5 stars are labeled as positive (1) and all others as negative (0). Table \ref{tab:amazon_reviews_selected_lang} details the specific splits.
    
    
    
    \item \text{ScienceQA:} A multimodal, multi-class question-answering dataset. Each entry consists of a question, multiple-choice answers, context (lecture, hint), and an optional image. The model must process both the unified text and the image (if present) to select the correct answer index. This dataset tests the model's ability to handle mixed-modality byte streams.
\end{itemize}

Table \ref{tab:dataset_cards} provides further details of the datasets. 


\begin{table}[t!]
    \centering
    \caption{Amazon Reviews Dataset (3 Languages).}
    \label{tab:amazon_reviews_selected_lang}
    \resizebox{\columnwidth}{!}{
    \begin{tabular}{lrrrr}
        \toprule
        \text{Language} & \text{\# Train} & \text{\# Val} & \text{\# Test} & \text{Pos/Neg Ratio} \\
        \midrule
        English & 200,000 & 5,000 & 5,000 & 40.0\% / 60.0\% \\
        German  & 200,000 & 5,000 & 5,000 & 40.0\% / 60.0\% \\
        Chinese & 200,000 & 5,000 & 5,000 & 40.0\% / 60.0\% \\
        \bottomrule
    \end{tabular}
    }
\end{table}

\subsection{Evaluation Metrics}
\label{sec:metrics}

\subsubsection{Model Performance Metrics}

\vspace{6pt}

\begin{itemize}
    \item Accuracy: The percentage of correctly classified samples, calculated as $\frac{\text{Correct Predictions}}{\text{Total Predictions}}$.
    
    \item F1-Score (Weighted): The weighted average of the F1-score for each class, accounting for label imbalance. This provides a robust measure of classification performance beyond simple accuracy.
    
    \item AUC (Area Under the ROC Curve): The Area Under the Receiver Operating Characteristic curve (i.e., the false positive/true positive trade-off for various values of the decision threshold). For binary tasks, this is the standard AUC. For multi-class tasks, we report the one-vs-rest (OvR) weighted AUC, which evaluates the model's ability to discriminate between each class and the rest.
\end{itemize}

\subsubsection{Resource \& Efficiency Metrics}


In addition to the accuracy, in a federated learning environment it is fundamental to measure the computational efficiency of an approach. For this, we employ the following metrics:

\begin{itemize}
    \item Model Parameters (M): The total count of all trainable parameters in the model, reported in millions. This is calculated by summing `p.numel()' for all parameters `p' in the model.
    
    \item Model Size (GB): The disk footprint of the model, calculated as the sum of memory required for all parameters $(\sum\text{p.nelement}() \times \text{p.element\_size}())$, reported in Gigabyte (GB).
    
    \item Total Time (minutes): The total wall-clock time in minutes for a single experiment run, including all training epochs and the final evaluation.
    
    \item GPU and CPU RAM Used (GB): The peak increase in system GPU and CPU RAM (in Gigabytes) observed during the model's training and evaluation. This is measured by taking a snapshot of `psutil.virtual\_memory().used' before and after the run.
    
   
    
    
    \item Network Footprint (MB):    
    A simulated metric for federated learning overhead. We report Total Sent and Total Received (S/R), where each is calculated as:
    $ \text{Total Sent/Received} = N_{\text{clients\_per\_round}} \times \text{Model Size (MB)} \times N_{\text{rounds}}$.
\end{itemize}

\subsection{Baseline Models}
\label{sec:baselines}

We compare our proposed models against three comparable baselines representing token-level, byte-level, and recurrent sequential approaches:

\begin{itemize}
    \item GPT-2 Medium: A standard GPT-2 model with a classification head. It operates on BPE-tokenized text, with a maximum sequence length of 1024 tokens. We adopted this model as an instance of small-sized transformer models that employs conventional tokenization.
    
    \item ByT5: The `google/byt5-small' model with no tokenizer with a classification head. This model operates directly on UTF-8 bytes and serves as a powerful byte-level baseline.
    
    \item LSTM: A 2-layer LSTM model with a byte-level embedding (257 tokens, including a pad token). It processes sequences of up to 1024 bytes.
\end{itemize}

Table \ref{tab:model_size_comparison_appendix} provides a comparison of the trainable parameter counts for the models evaluated in our experiments.

\begin{table}[h]
    \centering
    \caption{Model Size (Number of Trainable Parameters).}
    \label{tab:model_size_comparison_appendix}
    \resizebox{\columnwidth}{!}{
    \begin{tabular}{lrr}
        \toprule
        \text{Model} & \text{Hidden Dim} & \text{Param Count} \\
        \midrule
        ByT5 (small)   & 512   & 300M \\
        GPT-2 Medium   & 768   & 124M \\
        LSTM (2-layer) & 512   & 4.3M (approx.) \\
        FederatedBytes-Base & 512   & 18.6 (approx.) \\
        FederatedBytes-Performer-30M & 512   & 30M (approx.) \\
        FederatedBytes-Performer-100M & 512   & 100M (approx.) \\
        \bottomrule
    \end{tabular}
    }
\end{table}

\subsection{Proposed Model}
\label{sec:proposed}


The proposed model leverages a hierarchical patch-based embedding system which permits much longer contexts (up to 16,384 bytes) with comparable or smaller overall model size. We have implemented it in two variants:

\begin{itemize}
    \item FederatedBytes-Base: This model uses a standard 2-layer `TransformerEncoder' to process the byte patches, and leverages a conventional quadratic attention.
    
    \item FederatedBytes-Performer: This model replaces the standard transformer encoder with a 4-layer `Performer' (using the FAVOR+ linear attention \cite{likhosherstov2022favor}) to more efficiently handle long patch sequences. We have implemented this model in two sizes, 30M and 100M parameters, respectively.
    
\end{itemize}


All the hyperparameters used for our experiments are detailed in Table \ref{tab:hyperparams} in the Appendix.

\section{Results}
\label{sec:results}

\begin{table*}[!htbp]
\centering
\caption{Centralized performance of single models on the Amazon and ScienceQA datasets. Note: The top AUC scores within the Byte-Level model category are highlighted in bold. Text-Level baselines updated with GPT-2 Medium results.}
\label{tab:performance_consolidated}
\resizebox{\textwidth}{!}{%
\begin{tabular}{@{}llrrccccc@{}}
\toprule
\text{Language} & \text{Model} & \text{Model Params} & \text{Model Size (GB)} & \text{Acc (\%)} & \text{F1 (\%)} & \text{AUC (\%)} & \text{Mem (CPU/GPU)} & \text{Time (min)} \\
\midrule
\multicolumn{9}{l}{\textbf{Byte-Level Models}} \\
\text{Amazon (en)} & FederatedBytes-Base & 18.6M & 0.07 & 64.00 & 60.26 & 65.78 & 0.00 GB / 1.56 GB & 2.84 \\
& FederatedBytes-Performer-30M & 32.5M & 0.12 & 73.60 & 73.76 & 79.68 & 0.00 GB / 0.69 GB & 5.53 \\
& FederatedBytes-Performer-100M & 101.3M & 0.38 & 70.40 & 70.56 & \textbf{80.50} & 0.05 GB / 1.96 GB & 7.67 \\
& ByT5 & 300.8M & 1.12 & 59.80 & 44.76 & 58.45 & 0.00 GB / 19.99 GB & 17.31 \\
& LSTM & 4.3M & 0.02 & 59.80 & 44.94 & 50.08 & 0.01 GB / 0.28 GB & 4.21 \\
\cmidrule{1-9}
\text{Amazon (de)} & FederatedBytes-Base & 18.6M & 0.07 & 65.00 & 64.54 & 67.99 & 0.02 GB / 1.56 GB & 2.83 \\
& FederatedBytes-Performer-30M & 32.5M & 0.12 & 72.50 & 72.76 & 80.71 & 0.00 GB / 0.69 GB & 5.54 \\
& FederatedBytes-Performer-100M & 101.3M & 0.38 & 75.50 & 75.68 & \textbf{82.79} & 0.00 GB / 1.96 GB & 7.66 \\
& ByT5 & 300.8M & 1.12 & 59.90 & 44.98 & 66.53 & 1.63 GB / 19.99 GB & 17.30 \\
& LSTM & 4.3M & 0.02 & 59.80 & 44.76 & 50.29 & 0.00 GB / 0.28 GB & 4.22 \\
\cmidrule{1-9}
\text{Amazon (zh)} & FederatedBytes-Base & 18.6M & 0.07 & 65.60 & 65.91 & 72.14 & 0.02 GB / 1.56 GB & 2.83 \\
& FederatedBytes-Performer-30M & 32.5M & 0.12 & 73.60 & 73.77 & 79.95 & 0.66 GB / 0.68 GB & 5.62 \\
& FederatedBytes-Performer-100M & 101.3M & 0.38 & 71.50 & 71.77 & \textbf{80.34} & 0.07 GB / 1.95 GB & 7.76 \\
& ByT5 & 300.8M & 1.12 & 59.80 & 44.76 & 69.29 & 0.15 GB / 19.99 GB & 17.29 \\
& LSTM & 4.3M & 0.02 & 59.80 & 44.76 & 50.08 & 0.06 GB / 0.28 GB & 4.19 \\
\cmidrule{1-9}
\text{ScienceQA} & FederatedBytes-Performer (38M) & 38.8M & 0.15 & 39.00 & 36.17 & \textbf{62.60} & 2.42 GB / 0.15 GB & 1.84 \\
& ByT5 & 299.6M & 1.14 & 26.00 & 24.77 & 53.40 & 0.26 GB / 3.37 GB & 13.60 \\
\midrule
\midrule
\multicolumn{9}{l}{\textbf{Token-Level Models} (upper-bound reference for text tasks)} \\
\text{Amazon (en)} & GPT-2 Medium & 355M & 1.32 & 88.80 & 88.78 & 94.84 & 0.8 GB / 15.16 GB & 20.81 \\
\text{Amazon (de)} & GPT-2 Medium & 355M & 1.32 & 80.17 & 80.28 & 88.76 & 0.8 GB / 15.16 GB & 20.39 \\
\text{Amazon (zh)} & GPT-2 Medium & 355M & 1.32 & 77.77 & 77.70 & 84.54 & 0.8 GB / 15.16 GB & 21.40 \\
\bottomrule
\end{tabular}%
}
\end{table*}

\begin{table*}[!htbp]
\centering
\caption{Federated learning results comparing FederatedBytes-Performer-30M, FederatedBytes-Performer-100M, and ByT5. The highest AUC for each comparison is bolded. CPU memory reflects the active process usage (total) rather than differential deltas.}
\label{tab:FederatedBytes_network_scaling}
\resizebox{\textwidth}{!}{%
\begin{tabular}{@{}llrrrrrrrrrr@{}}
\toprule
\textbf{Dataset} & \textbf{Model} & \textbf{Clients} & \textbf{Acc (\%)} & \textbf{F1 (\%)} & \textbf{AUC (\%)} & \textbf{Time (min)} & \textbf{CPU (GB)} & \textbf{GPU (GB)} & \textbf{Net (GB)} & \textbf{Size (GB)} \\
\midrule
\multirow{15}{*}{\textbf{Amazon (en)}} 
& FederatedBytes-Performer-30M & 5 & 59.70 & 44.71 & \textbf{76.79} & 2.27 & 0.13 & 0.46 & 2.27 & 0.11 \\
& FederatedBytes-Performer-100M & 5 & 59.80 & 44.76 & 74.80 & 7.55 & 0.17 & 0.72 & 7.55 & 0.38 \\
& ByT5 & 5 & 59.80 & 44.75 & 60.61 & 22.32 & 0.28 & 1.71 & 22.32 & 1.12 \\
\cmidrule{2-11}
& FederatedBytes-Performer-30M & 10 & 59.80 & 44.76 & \textbf{72.04} & 5.67 & 0.13 & 0.46 & 5.67 & 0.11 \\
& FederatedBytes-Performer-100M & 10 & 59.80 & 44.75 & 72.04 & 18.87 & 0.17 & 0.72 & 11.32 & 0.38 \\
& ByT5 & 10 & 59.80 & 44.76 & 56.54 & 55.81 & 0.27 & 1.71 & 55.81 & 1.12 \\
\cmidrule{2-11}
& FederatedBytes-Performer-30M & 20 & 59.80 & 44.76 & 66.70 & 11.33 & 0.13 & 0.46 & 11.33 & 0.11 \\
& FederatedBytes-Performer-100M & 20 & 59.80 & 44.76 & \textbf{68.21} & 37.74 & 0.17 & 0.72 & 37.74 & 0.38 \\
& ByT5 & 20 & 59.80 & 44.75 & 49.66 & 111.62 & 0.27 & 1.71 & 111.62 & 1.12 \\
\cmidrule{2-11}
& FederatedBytes-Performer-30M & 50 & 40.20 & 44.75 & \textbf{64.95} & 28.33 & 0.13 & 0.46 & 28.33 & 0.11 \\
& FederatedBytes-Performer-100M & 50 & 59.80 & 44.75 & 63.87 & 94.34 & 0.17 & 0.72 & 94.34 & 0.38 \\
& ByT5 & 50 & 59.80 & 44.76 & 50.70 & 279.06 & 0.27 & 1.71 & 279.06 & 1.12 \\
\cmidrule{2-11}
& FederatedBytes-Performer-30M & 100 & 59.80 & 44.76 & \textbf{65.96} & 56.67 & 0.13 & 0.46 & 56.67 & 0.11 \\
& FederatedBytes-Performer-100M & 100 & 59.80 & 44.75 & 59.32 & 188.69 & 0.17 & 0.72 & 188.69 & 0.38 \\
& ByT5 & 100 & 59.7 & 45.24 & 51.24 & 558.12 & 0.27 & 1.71 & 558.12 & 1.12 \\
\midrule
\multirow{15}{*}{\textbf{Amazon (de)}} 
& FederatedBytes-Performer-30M & 5 & 63.40 & 52.93 & \textbf{80.25} & 2.27 & 0.13 & 0.46 & 2.27 & 0.11 \\
& FederatedBytes-Performer-100M & 5 & 59.80 & 44.76 & 79.77 & 7.55 & 0.17 & 0.72 & 7.55 & 0.38 \\
& ByT5 & 5 & 62.30 & 50.90 & 65.53 & 22.32 & 0.28 & 1.71 & 22.32 & 1.12 \\
\cmidrule{2-11}
& FederatedBytes-Performer-30M & 10 & 59.80 & 44.76 & \textbf{77.65} & 5.67 & 0.13 & 0.46 & 5.67 & 0.11 \\
& FederatedBytes-Performer-100M & 10 & 45.90 & 35.18 & 75.40 & 18.87 & 0.17 & 0.72 & 18.87 & 0.38 \\
& ByT5 & 10 & 59.80 & 44.76 & 60.84 & 55.81 & 0.27 & 1.71 & 55.81 & 1.12 \\
\cmidrule{2-11}
& FederatedBytes-Performer-30M & 20 & 66.40 & 59.59 & 74.05 & 11.33 & 0.13 & 0.46 & 11.33 & 0.11 \\
& FederatedBytes-Performer-100M & 20 & 59.80 & 44.76 & \textbf{75.86} & 37.74 & 0.17 & 0.72 & 37.74 & 0.38 \\
& ByT5 & 20 & 59.80 & 44.76 & 54.94 & 111.62 & 0.27 & 1.71 & 111.62 & 1.12 \\
\cmidrule{2-11}
& FederatedBytes-Performer-30M & 50 & 60.70 & 47.59 & \textbf{68.14} & 28.33 & 0.13 & 0.46 & 28.33 & 0.11 \\
& FederatedBytes-Performer-100M & 50 & 59.80 & 44.76 & 63.78 & 94.34 & 0.17 & 0.72 & 94.34 & 0.38 \\
& ByT5 & 50 & 59.80 & 44.76 & 51.77 & 279.06 & 0.28 & 1.71 & 279.06 & 1.12 \\
\cmidrule{2-11}
& FederatedBytes-Performer-30M & 100 & 59.80 & 44.76 & \textbf{70.43} & 56.67 & 0.13 & 0.46 & 56.67 & 0.11 \\
& FederatedBytes-Performer-100M & 100 & 44.90 & 34.11 & 66.71 & 188.69 & 0.17 & 0.72 & 188.69 & 0.38 \\
& ByT5 & 100 & 59.80 & 44.76 & 52.20 & 558.12 & 0.28 & 1.71 & 558.12 & 1.12 \\
\midrule
\multirow{15}{*}{\textbf{Amazon (zh)}} 
& FederatedBytes-Performer-30M & 5 & 61.60 & 49.02 & \textbf{79.56} & 2.27 & 0.13 & 0.46 & 2.27 & 0.11 \\
& FederatedBytes-Performer-100M & 5 & 59.80 & 44.75 & 75.15 & 7.55 & 0.17 & 0.72 & 7.55 & 0.38 \\
& ByT5 & 5 & 59.90 & 44.98 & 66.84 & 22.32 & 0.28 & 1.71 & 22.32 & 1.12 \\
\cmidrule{2-11}
& FederatedBytes-Performer-30M & 10 & 59.80 & 44.76 & \textbf{77.15} & 5.67 & 0.14 & 0.46 & 5.67 & 0.11 \\
& FederatedBytes-Performer-100M & 10 & 63.7 & 54.00 & 76.69 & 18.87 & 0.17 & 0.72 & 15.10 & 0.38 \\
& ByT5 & 10 & 59.80 & 44.75 & 59.72 & 55.81 & 0.27 & 1.71 & 55.81 & 1.12 \\
\cmidrule{2-11}
& FederatedBytes-Performer-30M & 20 & 40.10 & 44.75 & 70.65 & 11.33 & 0.13 & 0.46 & 11.33 & 0.11 \\
& FederatedBytes-Performer-100M & 20 & 64.20 & 54.93 & \textbf{75.83} & 37.74 & 0.17 & 0.72 & 30.19 & 0.38 \\
& ByT5 & 20 & 59.80 & 44.76 & 51.55 & 111.62 & 0.28 & 1.71 & 111.62 & 1.12 \\
\cmidrule{2-11}
& FederatedBytes-Performer-30M & 50 & 40.10 & 44.75 & 73.86 & 28.33 & 0.14 & 0.46 & 28.33 & 0.11 \\
& FederatedBytes-Performer-100M & 50 & 50.10 & 43.71 & \textbf{76.12} & 94.34 & 0.17 & 0.72 & 37.74 & 0.38 \\
& ByT5 & 50 & 59.80 & 44.75 & 50.38 & 279.06 & 0.28 & 1.71 & 279.06 & 1.12 \\
\cmidrule{2-11}
& FederatedBytes-Performer-30M & 100 & 40.20 & 44.75 & \textbf{72.87} & 56.67 & 0.14 & 0.46 & 56.67 & 0.11 \\
& FederatedBytes-Performer-100M & 100 & 59.80 & 44.76 & 68.20 & 188.69 & 0.17 & 0.72 & 188.69 & 0.38 \\
& ByT5 & 100 & 59.80 & 44.76 & 49.72 & 558.12 & 0.28 & 1.71 & 558.12 & 1.12 \\
\bottomrule
\end{tabular}%
}
\end{table*}

\begin{figure*}[!htbp]
    \centering
    \includegraphics[width=\textwidth]{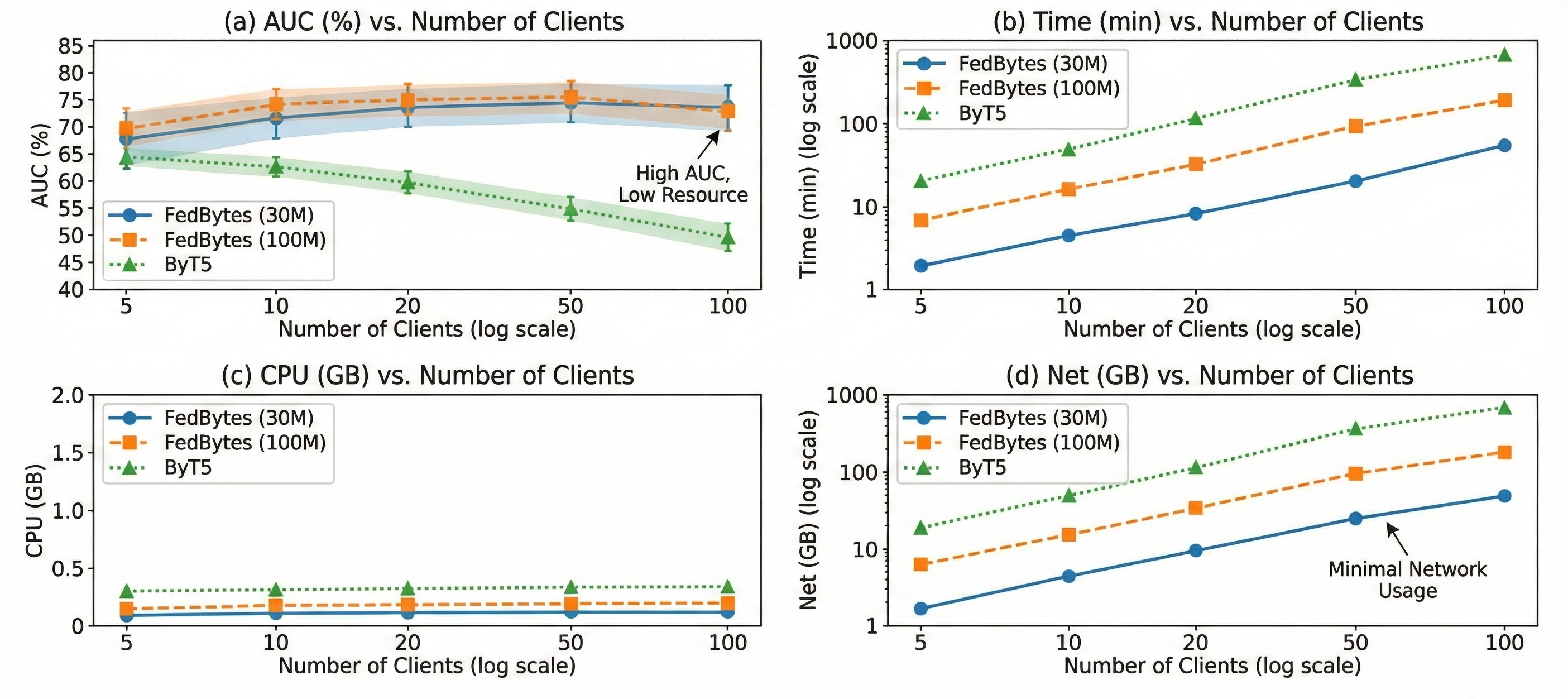}
    \caption{\textbf{FederatedBytes Scalability and Efficiency Analysis.} 
    \textbf{Top:} FederatedBytes maintains high AUC scores while ByT5 collapses as client counts increase. 
    \textbf{Bottom:} FederatedBytes demonstrates significantly lower computational overhead, network traffic, and memory footprint compared to ByT5. FederatedBytes-Performer-30M demonstrates superior efficiency, maintaining competitive AUC scores with significantly lower training time, CPU memory, and network consumption compared to the larger ByT5 model, especially as the number of clients increases. Data points represent averages across Amazon (en), (zh), and (de) datasets.}
    \label{fig:combined_analysis}
\end{figure*}

In this section, we present the results of our centralized and federated experiments, focusing on the trade-offs between token- and byte-level processing.

\subsection{Centralized Results}
In the experiments carries out in the centralized setting, all models were trained as single models over the full training sets. Table \ref{tab:performance_consolidated} reports the results, which include metrics to evaluate both the effectiveness and the efficiency of the compared models.





To begin with, Table \ref{tab:performance_consolidated} shows that the conventional token-level GPT-2 model has been able to achieve the strongest accuracy metrics on all the text-only tasks (Amazon (en,de,zh)). However, this model is approximately 3.5$\times$ larger than the largest of our models, FederatedBytes-Performer-100M, has occupied $\sim$8$\times$ the GPU memory (15.6 GB vs 1.96 GB) and has proved $\sim$3$\times$ slower to run (20-21 minutes vs 7.7). By contrast, our FederatedBytes-Performer has still been able to achieve a remarkable performance on the text tasks, reaching 80.50 AUC pp on Amazon (en), with a much smaller computational footprint, while at the same time being able to process a multimedia dataset such as ScienceQA. FederatedBytes has also compared very favorably with ByT5, proving more accurate and more efficient in every metric. For instance, on the Amazon (en) dataset FederatedBytes-Performer-100M has achieved an AUC of 80.50 pp vs 58.45 pp for ByT5, in less than half the runtime and with 1/10 of the GPU memory footprint. For the ScienceQA dataset, only ByT5 and FederatedBytes\footnote{Note that we have used a 38M parameter version for this task, that corresponds to the 30M text-alone version augmented with the image input.} have been able to perform the task in the first place, and FederatedBytes outperformed By5 in every metric. At their turn, the baseline LSTM and the FederatedBytes-Base model have struggled to match the performance of the more complex models.

\subsection{Federated Learning Results}



\label{subsec:fl_results}

We benchmarked the scalability and robustness of FederatedBytes (30M and 100M variants) against the ByT5 baseline in a simulated federated learning setting on the Amazon (en, de, zh) datasets. The experiments ranged from 5 to 100 participating clients to evaluate performance under increasing model fragmentation.

As shown in Table \ref{tab:FederatedBytes_network_scaling}, FederatedBytes has demonstrated a superior scalability:
\begin{itemize}

    \item \textbf{Resistance to Model Collapse:} In high-concurrency settings (100 clients), the ByT5 baseline has consistently collapsed, with AUC dropping to near-random performance ($\approx 50-52\%$). In contrast, FederatedBytes-Performer-30M has maintained robust learning, achieving AUC scores of 65.96\% (en), 72.87\% (zh), and 66.71\% (de). Detailed results are shown in Figure \ref{fig:combined_analysis}.
    
    \item \textbf{Efficiency Gains:} FederatedBytes-Performer-30M has reduced training time by approximately 10$\times$ compared to ByT5 (e.g., 56 min vs. 558 min for 100 clients), while consuming significantly less CPU and network bandwidth and maintaining a peak memory footprint of only 152 MB compared to ByT5's 3.4 GB. 
    
    \item \textbf{Generalization:} Interestingly, the federated FederatedBytes-Performer model frequently has outperformed its centralized counterpart (e.g., 78.8\% AUC vs 71.9\% for the centralized model on Amazon (en)), suggesting that the federated averaging process acts as an effective regularizer for our byte-level architecture.

    \item \textbf{Multimodal Performance:} FederatedBytes-Performer has displayed interesting multimodal capabilities on ScienceQA, reporting an AUC of 62.60\% vs. ByT5's 53.40\%.
    
\end{itemize}

We also observe that the 30M parameter model has achieved a higher AUC than the 100M model as the number of clients increases. This occurs because dividing the data among many clients leaves each client with an increasingly small dataset ($|D_{local}|$), and the large 100M model is too complex for such small data ($|\theta| \gg |D_{local}|$) and ends up ``memorizing'' local noise (overfitting). When these noisy updates are then combined, they conflict and degrade the global model. In contrast, the 30M model has fewer parameters which helps prevent it from memorizing noise and learn more stable, general patterns that improve the overall accuracy.

\section{Conclusion}
\label{sec:conclusion}

This paper has proposed FederatedBytes, the first byte-level transformer model expressly engineered for federated learning environments. Byte-level models are particularly attractive for federated learning as they are able to dispense with memory-occupying language-specific vocabularies, remove the ``vocabulary leak'' privacy gap, and process the diverse type of data that may originate from different clients. By adopting the memory-sparing Performer attention, we have been able to mitigate the quadratics cost of standard self-attention, permitting efficient processing on resource-constrained edge devices.



Looking ahead, we aim to extend FederatedBytes with deeper multimodal capabilities, scale federated training to larger client populations and longer contexts, and incorporate personalized optimization methods to better accommodate diverse user data, all the while preserving privacy and communication efficiency.

\section{Acknowledgment}

The authors are grateful to the Food Agility Cooperative Research Centre and Robert Bosch (Australia) Pty Ltd for their generous support toward this research.
This project was supported by funding from Food Agility CRC Ltd, funded under the Commonwealth Government CRC Program. The CRC Program supports industry-led collaborations between industry, researchers and the community.



\bibliographystyle{elsarticle-harv}


\appendix

\section{Hyperparameters}

The main hyperparameters for the proposed model are reported in Table \ref{tab:hyperparams}.

\begin{table}[h!]
\centering
\caption{Key hyperparameters and configuration settings used in the experiments.}
\label{tab:hyperparams}
\resizebox{0.8\columnwidth}{!}{
\begin{tabular}{@{}ll@{}}
\toprule
\textbf{Parameter} & \textbf{Value} \\ \midrule
\multicolumn{2}{l}{\textit{General Optimization}} \\
Random Seed & 42 \\
Optimizer & AdamW \\
Learning Rate & $3.0 \times 10^{-5}$ \\
Weight Decay & 0.01 \\
Dropout & 0.1 \\
\midrule
\multicolumn{2}{l}{\textit{Training Dynamics}} \\
Micro-Batch Size ($B$) & 2 \\
Gradient Accumulation Steps & 128 \\
Effective Batch Size & 256 \\
Centralized Epochs & 15 \\
\midrule
\multicolumn{2}{l}{\textit{Federated Learning (Simulation)}} \\
FL Strategy & FedAvg \\
Total Clients ($K$) & \{5, 20, 50, 100\} \\
Client Fraction (Fit) & 0.5 \\
Local Epochs ($E$) & 5 \\
FL Rounds ($T$) & 5 \\
\midrule
\multicolumn{2}{l}{\textit{Model Architecture (FederatedBytes)}} \\
Hidden Size ($d$) & 768 \\
Attention Heads & 12 \\
Encoder Layers & 12 \\
Max Sequence Length ($L$) & 16,384 bytes \\
Patch Size ($P_s$) & 128 bytes \\
Hierarchical Encoder & Conv1d + Pool \\
Attention Type & Performer (ORFs) \\ \bottomrule
\end{tabular}
}
\end{table}

\section{GPT-2 Model Configuration and Model Card}
\label{sec:gpt2_model_card}
Our GPT-2 baseline follows the HuggingFace \texttt{gpt2} recipe:
\begin{itemize}
    \item \text{Number of Layers:} 12
    \item \text{Hidden Size (d\_model):} 768
    \item \text{Number of Attention Heads:} 12
    \item \text{Intermediate Feed-Forward Size:} 3072
    \item \text{Vocabulary Size:} 50,257
    \item \text{Context window:} 1024 tokens
\end{itemize}

\noindent \text{Model Card:} In alignment with \cite{wolf2020huggingfaces}, we provide brief model documentation:
\begin{description}
    \item[Use Cases:] General-purpose language modeling, next-token prediction, and fine-tuning for classification tasks.
    \item[Limitations:] Prone to generate plausible-sounding but factually incorrect or biased text; context window limited to 1024 tokens in the base model.
    \item[License:] Licensed under the MIT or Apache-2.0 (check original GPT-2 license).
\end{description}

\section{Federated Training Algorithm}

Algorithm~\ref{alg:federated_training} outlines the overall federated training process.

\begin{algorithm}[t]
\caption{FederatedBytes Training Procedure (Simulation)}
\label{alg:federated_training}
\begin{algorithmic}[1]
\State \textbf{Input:} Global model $\theta^0$, Learning rate $\eta$, Clients $K$, Rounds $T$, Local Epochs $E$, Micro-batch $B$, Accumulation Steps $M$, Patch size $P_s$
\State \textbf{Output:} Final global model $\theta^T$

\For{round $t=0, \dots, T-1$}
    \State Server selects subset of clients $S_t \subseteq \{1, \dots, K\}$
    \State Server broadcasts $\theta^t$ to clients $k \in S_t$
    
    \For{each client $k \in S_t$ \textbf{in parallel}}
        \State $\theta_k \gets \theta^t$ \Comment{Initialize local model}
        \State $g_{\text{accum}} \gets 0$ \Comment{Reset gradient accumulator}
        
        \For{epoch $e=1, \dots, E$}
            \For{step $j, (X, Y)$ in $\mathcal{D}_k$ (Micro-batch size $B$)}
                \State \textit{// Step 1: Byte-Level Processing}
                \State $b \gets \text{UTF8Encode}(X)$
                \State $b_{\text{pad}} \gets \text{Pad}(b, L_{\text{max}})$ \Comment{Pad to 16,384 bytes}
                
                \State \textit{// Step 2: Patching \& Hierarchical Encoding}
                \State $P \gets \text{Segment}(b_{\text{pad}}, \text{size}=P_s)$ \Comment{Create 128-byte patches}
                \State $Z \gets \text{Embed}(P)$ \Comment{Map bytes to vectors}
                \State $Z_{\text{enc}} \gets \text{Pool}(\sigma(\text{Conv1d}(Z)))$ \Comment{Hierarchical Encode}
                
                \State \textit{// Step 3: Efficient Encoding}
                \State $H \gets \text{PerformerAttention}(Z_{\text{enc}})$ \Comment{Linear $O(L)$ Attention}
                \State $\hat{Y} \gets \text{Classifier}(\text{MeanPool}(H))$
                
                \State \textit{// Step 4: Optimization with Accumulation}
                \State $\mathcal{L} \gets \text{CrossEntropy}(\hat{Y}, Y)$
                \State $g_{\text{accum}} \gets g_{\text{accum}} + \frac{1}{M} \nabla_{\theta_k} \mathcal{L}$ \Comment{Accumulate Gradients}
                
                \If{$(j \mod M == 0)$ or (end of data)}
                    \State $\theta_k \gets \theta_k - \eta \cdot \text{AdamW}(g_{\text{accum}})$
                    \State $g_{\text{accum}} \gets 0$ \Comment{Reset accumulator}
                \EndIf
            \EndFor
        \EndFor
        \State \text{Client} sends $\theta_k$ to Server
    \EndFor
    
    \State \textit{// Server Aggregation (FedAvg)}
    \State $\theta^{t+1} \gets \sum_{k \in S_t} \frac{N_k}{\sum N_j} \theta_k$
\EndFor
\end{algorithmic}
\end{algorithm}

\end{document}